\pdfoutput=1
\documentclass[conference]{IEEEtran}
\IEEEoverridecommandlockouts
\usepackage{cite}
\usepackage{amsmath,amssymb,amsfonts}
\usepackage{algorithmic}
\usepackage{graphicx}
\usepackage{textcomp}
\usepackage{xcolor}
\usepackage{comment}
\usepackage{makecell}
\usepackage{tablefootnote}
\usepackage{float}
\usepackage{footnote}
\usepackage{booktabs}
\usepackage{tabu}
\usepackage{multirow}
\usepackage[super]{nth}
\usepackage[ colorlinks = true,
  anchorcolor = blue,
  filecolor = blue,
  citecolor=blue,
  linkcolor=blue,
  urlcolor=blue]{hyperref}
\usepackage{makecell}
\usepackage[utf8x]{inputenc}

\def\BibTeX{{\rm B\kern-.05em{\sc i\kern-.025em b}\kern-.08em
    T\kern-.1667em\lower.7ex\hbox{E}\kern-.125emX}}
  
\begin{document}

\makeatletter
\newcommand{\linebreakand}{%
  \end{@IEEEauthorhalign}
  \hfill\mbox{}\par
  \mbox{}\hfill\begin{@IEEEauthorhalign}
}

\title{The Effect of Normalization for Bi-directional Amharic-English Neural Machine Translation\\
}
\author{\IEEEauthorblockN{Tadesse Destaw Belay}
\IEEEauthorblockA{\textit{College of Informatics} \\
\textit{Wollo University}\\
Kombolcha, Ethiopia \\
tadesseit@gmail.com}
\and
\IEEEauthorblockN{Atnafu Lambebo Tonja}
\IEEEauthorblockA{\textit{ Centro de Investigación en Computación} \\
\textit{Instituto Politécnico Nacional}\\
Mexico City, Mexico \\
alabedot2022@cic.ipn.mx}
\and
\IEEEauthorblockN{Olga Kolesnikova}
\IEEEauthorblockA{\textit{ Centro de Investigación en Computación} \\
\textit{Instituto Politécnico Nacional}\\
Mexico City, Mexico \\
kolesolga@gmail.com
}
\and
\IEEEauthorblockN{Seid Muhie Yimam}
\IEEEauthorblockA{\textit{Dept. of Informatics} \\
\textit{Universität Hamburg}\\
Hamburg, Germany \\
seid.muhie.yimam@uni-hamburg.de}
\and
\IEEEauthorblockN{Abinew Ali Ayele}
\IEEEauthorblockA{\textit{ICT4D Research Center} \\
\textit{Bahir Dar University}\\
Bahir dar, Ethiopia \\
abinewaliayele@gmail.com}
\and
\IEEEauthorblockN{Silesh Bogale Haile}
\IEEEauthorblockA{\textit{Dept. of  Computer Science} \\
\textit{Assosa University}\\
Assosa, Ethiopia \\
sileshibogale123@gmail.com}

\and
\linebreakand

\IEEEauthorblockN{Grigori Sidorov}
\IEEEauthorblockA{\textit{ Centro de Investigación en Computación} \\
\textit{Instituto Politécnico Nacional}\\
Mexico City, Mexico \\
sidorov@cic.ipn.mx }
\and
\IEEEauthorblockN{Alexander Gelbukh}
\IEEEauthorblockA{\textit{ Centro de Investigación en Computación} \\
\textit{Instituto Politécnico Nacional}\\
Mexico City, Mexico \\
gelbukh@cic.ipn.mx}
}
\maketitle
\begin{abstract}
Machine translation (MT) is one of the prominent tasks in natural language processing  whose objective is to translate texts automatically from one natural language to another. Nowadays, using deep neural networks for MT task has received a great deal of attention. These networks require lots of data to learn abstract representations of the input and store it in continuous vectors. This paper presents the first relatively large-scale Amharic-English parallel sentence dataset. Using these compiled data, we build bi-directional Amharic-English translation models by fine-tuning the existing Facebook M2M100 pre-trained model achieving a BLEU score of 37.79 in Amharic-English translation and 32.74  in English-Amharic translation. Additionally, we explore the effects of Amharic homophone normalization on the machine translation task. The results show that  normalization of Amharic homophone characters increases the performance of Amharic-English machine translation in both directions.
\end{abstract}


\section{Introduction}
Machine translation (MT) is a  sub-field of natural language processing (NLP) that investigates how to use computer software to automatically translate text or speech from one language to another without human involvement. 
MT is one of the prominent tasks in NLP that is tackled in several ways \cite{bojar2017findings}. The first MT research began at about 1950s and in 1952 the first International Conference on Machine Translation was organized at the Massachusetts Institute of Technology (MIT). It has long research history and experienced four stages, namely, Rule-based MT \cite{forcada2011apertium},  Statistical MT (SMT) \cite{koehn2007moses}, hybrid MT, and Neural MT (NMT) \cite{cho2014properties,kalchbrenner2013recurrent}.

The most severe drawback of the rule-based method is that it
has ignored the need for context information in the translation
process. It is highly dependent on hand-crafted features. Phrase-based SMT (PBSMT), the most prevalent version of SMT, generates translation by segmenting the source sentence into several phrases and performing phrase translation and replacement. It may ignore the long sentence dependency and require high computing devices \cite{baniata2021transformer}. Recently, using deep neural networks for MT task has received great attention. NMT also improves training procedures due to the end-to-end procedure without tedious feature engineering and complex setups. NMT employs such techniques as recurrent neural network (RNN) \cite{bahdanau2014neural}, convolutional neural network (CNN) \cite{gehring2016convolutional}, and self-attention network (Transformer) \cite{vaswani2017attention}.

Transformer models with the pre-training approach is a new NMT strategy entirely based on attention mechanisms proposed in 2017 \cite{vaswani2017attention}. Among the different neural network architectures, the Transformer model has emerged as the dominant NMT paradigm \cite{raganato2018analysis,yigezu2021multilingual, tonja2021parallel}. It has become the state-of-the-art model for many artificial intelligence tasks, including machine translation. In terms of model, the Transformer-based pre-trained models are fast to fine-tune, highly accurate and has been proven to outperform widely used recurrent networks \cite{zhang2018improving,baniata2021transformer,yang2020survey}. 

The focus of MT research for the Amharic language has been on rule-based and SMT methods. In this work, we used the transformer model as a baseline translation system to explore the applicability of Facebook M2M100 multi-lingual pre-trained language model for Amharic-English translation in both directions. Furthermore, this research work investigated the impact of normalization of the Amharic homophones on Amharic-English MT tasks.

The main contributions of this work are:
\begin{enumerate}
    \item Exploration of the Amharic-English and English-Amharic machine translation tasks.
    \item Introduction of the first large-scale publicly available Amharic-English translation parallel dataset.
    \item Development and implementation of state-of-the-art Amharic-English translation models.
    \item Investigation of the effect of Amharic homophone character normalization on the machine translation task.
\end{enumerate}

The rest of this paper is organized as follows. Section \ref{amharic_langauge} presents a detail description of Amharic language while Section \ref{motivation} shows the motivation for this research. In Section \ref{related_work}, we review related work. Section \ref{dataset} describes the existing parallel corpus and the collection of a new corpus from the news domain. The general pre-processing steps applied to both corpora are presented in Section \ref{data_pre}. Section \ref{models} discusses the proposed NMT models and Section \ref{experiment} gives the experimental results. In the end, Section \ref{conclusion} concludes the paper and sheds some light on possible future work.

\section{Amharic language}\label{amharic_langauge} 
 Amharic is the second most spoken Semitic language next to Arabic which has its own alphabet and writing scripts called 'Fidel', that was borrowed from Ge'ez, another Ethiopian Semitic language.  Fidel is a syllable-based writing system where the consonants and vowels co-exist within each graphic symbol. The Amharic language is spoken by more than 57 million people with up to 32 million native speakers and 25 million non-native speakers \cite{eberhard2020ethnologue}. Amharic is the working language of the Federal Democratic Republic of Ethiopia (FDRE) and for many regional states in the country. In Amharic, there are 34 core characters each having seven different derivatives to represent vowels. In addition, it has 20 labialized characters, more than 20 numerals, and 8 punctuation marks. Amharic uses a total of more than 310 characters. The language is known for being morphologically complex and it is highly inflectional. Unlike English, French, Spanish, Japanese, and Chinese, Amharic is considered low-resource because the data are not well organized and technologically less supported \cite{woldeyohannis2017experimenting}.

\section{Motivation}\label{motivation}
Nowadays advancement in technology has made the lifestyle of human beings much easier by helping daily activities. One of the applications that solved communication barriers between people speaking different languages is machine translation. Many big technology companies such as Google, Microsoft, IBM, etc. provide translation services for many languages to facilitate communication between people without using a human translator. However, the quality of NMT is massively dependent on quantity, quality, and relevance of the training dataset \cite{ahmadnia2019augmenting}. Such companies have achieved promising results for bilingual high-resource languages, but they are inadequate for low-resource languages like Amharic.

\begin{figure}[ht!]
    \centering
    \includegraphics[width=0.7\linewidth]{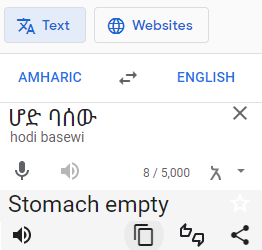}
   \caption{Examples of Google Amharic to English translation}
    \label{fig:motiv}
\end{figure}
Figure \ref{fig:motiv} shows Google translation of Amharic words into English. Google translated the Amharic input as "stomach empty"  which is a wrong translation. The correct literal meaning will be "he become Disappointed". This shows that the applications used by companies that provide translation system like Google require improvements. One of the cause for poor performance of MT systems for languages like Amharic is availability of limited resource in digital space \cite{ahmadnia2019augmenting}. In this research work we present a newly curated English-Amharic parallel dataset that can be used for MT research and help to solve the performance issue of an Amharic MT system.    
  
On the other hand, Amharic is one of morphologically rich language and normalization of the Amharic homophone characters might have an impact on such downstream NLP applications as MT and sentiment analysis \cite{belay2021impacts}. This research work is intended to study the effect of homophone normalization on Amharic-English machine translation. Furthermore, expanding the translation dataset and developing state-of-the-art bi-directional Amharic-English translation models are our motivations to carryout this research work.

\section{Related work} \label{related_work}
Many automatic translation works have been carried out for the major pairs of European and Asian languages, taking advantage of large-scale parallel corpora. However, very few research studies have been conducted on low-resource languages like Amharic to English due to its scarcity of parallel data. In this section, we have focused on exploring how machine translation is conducted for the Amharic language.
Among recent works, Biadgligne and Smaïli  \cite{biadgligne2021parallel} described the development of an English-Amharic Statistical Machine Translation (SMT) and Neural Machine Translation (NMT) experiments, achieving 26.47 and 32.44 BLEU scores, respectively. They harvested and used 225,304 parallel sentences in different domains. To the best of our knowledge, Biadgligne and Smaïli  \cite{biadgligne2021parallel} work is the largest dataset used in Amharic machine translation research work.  

Gezmu et al. \cite{gezmu2021extended} trained neural machine translation and phrase-based statistical machine translation models using 145,364 parallel English-Amharic sentences, achieved a BLEU score of 20.2 and 26.6 respectively. They concluded that neural machine translation models outperform phrase-based statistical machine translation models.
Abate et al. \cite{abate2018parallel} focused on the development of parallel corpora for English and five Ethiopian languages, including Amharic. They used 40,726 parallel sentences for Amharic and bi-directional Statistical Machine Translation (SMT) technique. Finally, they achieved a BLEU score 22.68  in Amharic-English translation. 

Teshome et al. \cite{teshome2015phoneme} considered the application of phone-based statistical machine translation for English-Amharic translation. They used 18,432 parallel sentences and achieved a BLEU score of 37.5, a gain of 2.21 BLEU point from their own previous baseline phrase-based MT experiments.
Further more, Tracey and Strassel \cite{tracey2020basic} developed a low-resource language dataset for emergent incidents (LORELEI). Their data includes multiple languages and was published in the Linguistic Data Consortium (LDC) where Amharic is among the languages with 60,884 Amharic-English parallel sentences. However, this dataset is not freely available for further experiments.

The remain and addressed related works are summarized in Table \ref{tab:table1}, with dataset and methods used. As we can see in this related work table, only a few studies have been conducted for the translation from Amharic into English or vice versa and most of them were conducted using traditional approaches with a small number of parallel sentences. This is due to unavailability of enough Amharic linguistic resources for deep learning experiments.

\begin{table*}[!ht]
\begin{center}
\caption{Amharic-English and English-Amharic MT studies in terms of dataset size,  method(s) used, and BLEU score achieved}
\label{tab:table1}
\begin{tabular}{@{}lllll@{}}
\hline
\textbf {Authors} & \textbf {Trans. direction}& \textbf {\# Dataset used} &  \textbf {Method(s)} & \textbf {BLEU score}\\ 
\hline
Biadgligne and Smaïli \cite{biadgligne2021parallel} & {En→Am} & 225,304 &   Statistical machine translation & 26.47 \\ 
& &  & Neural machine translation & 32.44 \\
\hline
Gezmu et al. \cite{gezmu2021extended} & Am→En & 45,364 & Phrase-based statistical machine translation  & 20.2 \\ 
& &  & Neural Machine Translation & 26.6 \\
\hline
Abate et al. \cite{abate2018parallel} & Am→En & 40,726 & Statistical machine translation & 22.68 \\ 
 &En→Am &  & Statistical machine translation & 13.31 \\
\hline
Teshome et al. \cite{teshome2015phoneme} & En→Am & 18,432 & Phoneme-based statistical machine translation & 37.5 \\ 
\hline
Teshome and Besacier \cite{teshome2012preliminary} & En→Am & 18,432 & Phrase-based statistical machine translation & 35.32 \\ 
\hline
Ashengo et al. \cite{ashengo2021context} & En→Am & 8,603 & Combination of context-based MT (CBMT) with RNN & 11.34 \\ 
\hline
Tadesse and Mekuria. \cite{tadesse2000} & En→Am & 37,970 & Statistical machine translation & 18.74\\ 
\hline  
Hadgu et al. \cite{hadgu2020evaluating} & Am→En & 977 & Google translate, Yandex translate  & 23.2, 4.8 \\ 
& En→Am & 1915 & Google translate, Yandex translate & 9.6, 1.3 \\
\hline
\end{tabular} 
\end{center}
\end{table*} 
\section{Building a Parallel Dataset} \label{dataset}
In this section, we have identified available data sources that we have used for the research work. As machine translation requires parallel documents as an input, Table \ref{tab:table2} shows the potential Amharic-English bi-lingual resources. It can be seen in the table that the largest parallel corpus for Amharic-English language pairs was collected by Biadgligne and Smaïli \cite{biadgligne:hal-03547539}. In addition to the available datasets in Table \ref{tab:table2}, we have contributed to the MT research field by creating a new parallel corpus with 33,955 sentence pairs extracted text from such news platforms as Ethiopian Press Agency\footnote{\href {https://www.press.et/} {https://www.press.et/}}, Fana Broadcasting Corporate\footnote{\href {https://www.fanabc.com/} {https://www.fanabc.com/}}, and  Walta Information Center\footnote{\href {https://waltainfo.com/} {https://waltainfo.com/}}. As the data we used is from different sources, it includes various domains such as religious (Bible and Quran), politics, economics, sports, news, among others. 

As one can see in Table \ref{tab:table2}, the total number of parallel sentences is about 1.1M, while the unique parallel sentences are 888,837. This is due to duplication in the sources we used. This unique parallel sentence is the largest to date then.
\section{Data pre-processing}\label{data_pre}
Before performing experiments, the development of every NLP task begins with a text pre-processing step \cite{kumar2012text}. As our data were collected from different sources, we noticed a lot of text irregularities. Some data was too noisy so we eliminated it from our corpus. Then we performed a series of pre-processing steps to canonize all tokens in Amharic and English sentences. In these steps, the following tasks were performed: data cleaning (removing emojis, URL), abbreviation expansion, Latin character lowercase, duplicated sentence removal, and Amharic homophone character normalization. For abbreviation expansion, we created a list of known abbreviations for both Amharic and English, then expanded them to their full writing form. Most of English abbreviations were collected from GitHub repositories\footnote{\url{https://github.com/JRC1995/Machine-Translation-Transformers}}. 
\begin{table}[H]
\centering
\caption{\label{tab:table2}Available Amharic-English parallel data sources}
\begin{tabular}{lcc}
\hline
\textbf{Data source} & \textbf{\# Sentence pairs} & \textbf{Accessible}\\
\hline
Am-En ELRA-W0074\tablefootnote{\url{ http://catalog.elra.info/en-us/repository/browse/ELRA-W0074/}}
 & 13,347 & yes\\
 
Biadgligne and Smaïli\cite{biadgligne2021parallel}
 & 225,304 & yes\\

Horn MT\tablefootnote{\url{https://github.com/asmelashteka/HornMT}} & 2,030 & yes\\

Am-En MT corpus\tablefootnote{\url{https://github.com/adtsegaye/Amharic-English-Machine-Translation-Corpus}} & 53,312 & yes\\

Gezmu et al.\cite{gezmu2021extended}
& 145,364 & yes\\

Abate et al.\cite{abate2018parallel}
& 40,726 & yes\\

Open Parallel Corpus (OPUS) \cite{lison2016opensubtitles2016} & 562,141 & yes\\

LORELEI-Amharic \cite{tracey2020basic}
& 60,884 & no\\ 

Admasethiopia\tablefootnote{\url{https://github.com/admasethiopia/parallel-text}} & 153 & yes\\

MT Evaluation Dataset\tablefootnote{\url{https://doi.org/10.5281/zenodo.3734260}} & 2,914 & yes\\

\textbf{Newly curated (our data)} & \textbf{33,955} & \textbf{yes}\\
\hline
\textbf{Total} & 1,140,130 & yes\\
\textbf{Unique sentence pairs} & 888,837 & yes\\
\hline
\end{tabular}
\end{table}
\textbf{Homophone character normalization}: 
In Amharic writing, there are different characters with the same sound which are called homophones. Homophones with different symbols in Amharic text might have different writing standards and different meanings. However, they are also considered as redundant alphabets by most of the users, specially by the online and social media communities. 

The current trend in Amharic NLP research is to normalize the homophone characters into a single representation \cite{woldeyohannis2017experimenting,abate2018parallel,gezmu2021extended,biadgligne2021parallel}. Belay et al. \cite{belay2021impacts} have studied the impact of normalization for some downstreamm NLP applications. They showed that homophone normalization improves the performance of some downstream applications such as information retrieval (IR) but not recommended for sentiment analysis. However, there is no study to show that normalization helps to build efficient machine translation models or not. We developed our translation models in the regular (unnormalized) and the normalized forms (representing different homophones as a single character) to analyze the impact of normalization. For the normalized model, we used frequency-based homophone normalizer \cite{belay2021impacts} replacing the set of characters with similar function with a single most frequently used character.

\begin{figure}[hbt!]
    \centering
    \includegraphics[width=\linewidth]{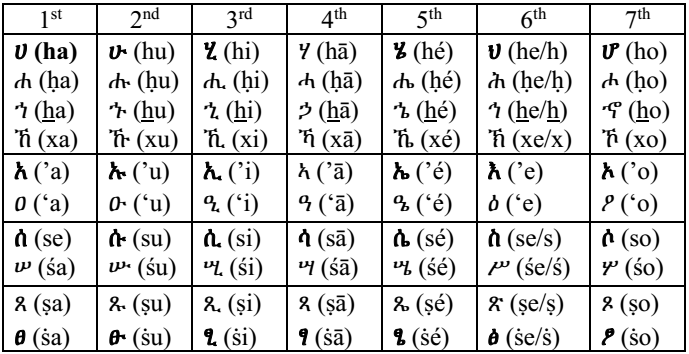}
   \caption{Amharic homophone characters with their 7 derivatives using their International Phonetic Alphabet (IPA) notation. The normalized homophones is the first one in boldface in cell of each row except the  of \nth{4} column of the \nth{1} and \nth{2} row where the \nth{1} character was used as the normalized one.}
    \label{fig:homophone}
\end{figure}


\section{The proposed neural machine translation models}\label{models}
In this section, we describe the transformer-based pre-trained models that could be used to train Neural Machine Translation systems.
Unlike other neural networks such as RNNs, the Transformer does not necessarily process the input data in sequential order. Instead, the self-attention mechanism identifies the context which gives meaning to each position in the input sequence, allowing more parallelization, and reducing the training time. The architecture of the Transformer network follows the so called encoder-decoder paradigm, trained in an end-to-end fashion. The encoder is used to represent the source sentence as a semantic vector, while the decoder takes in the semantic vector and makes prediction to a target sentence. 
The Transformer model, which applies a self-attention approach to measure the strength of a relationship between two words in a sentence (contextual information), has contributed to improving performance in MT and various natural language processing tasks. 


\subsection{Pre-trained language models (PLMs)}
Pre-trained Language Models (PLMs) are large neural networks used in a wide variety of NLP tasks \cite{elazar2021measuring}. Models are first pre-trained over a large text corpus and then fine-tuned on a downstream task. PLMs are thought of as good language encoders, supplying basic language understanding capabilities that can be used with ease for many downstream tasks \cite{elazar2021measuring,weng2020acquiring}.

Fan et al. \cite{fan2020beyond}, Facebook researchers, proposed a multilingual encoder-decoder model trained for Many-to-Many multilingual translation with 418M parameters (M2M100 418M). This multilingual machine translation model is based on the Transformer sequence-to-sequence architecture. The model can directly translate between the 9,900 directions of 100 languages. We used this pre-trained model for our bi-directional translation experiments. This pre-trained multi-lingual model is available at Hugging Face\footnote{\url{https://huggingface.co/facebook/m2m100_418M}}

\section{Experimental setup and results}\label{experiment}
\subsection{Experimental setup}
To demonstrate the effectiveness of our approach, we built our baseline Transformer models and fine-tuned the available pre-trained model. We used Google colab pro+ to train our bi-directional Amharic to English translation models. During model training, the parallel sentences for Amharic and English were divided into 80\% for training, 10\% validation, and 10\% for testing. Automatic evaluation was made using Bilingual Evaluation Understudy (BLEU) metric \cite{papineni2002bleu}. BLEU score is defined in the range between 0 and 1, where 1 is a perfect match with the reference and 0 is for no words matched. 

\textbf{Transformer baseline:} We trained Transformer sequence-to-sequence models from scratch for bi-directional Amharic-English NMT using OpenNMT with TensorFlow deep learning framework \cite{klein2017opennmt}. We tokenized the text using Byte Pair Encoding (BPE) \cite{gage1994new} subword tokenization, which is a simple form of data compression algorithm in which the most common pair of consecutive bytes of data is replaced with a byte that does not occur in that data. The parameters used to train the model are 512 hidden units, 6 layers, a learning rate of 0.0001, and a maximum step of 50K, a batch size of 32, and the Adam optimizer. 

\textbf{Pre-trained model:} We used the multilingual Facebook M2M-100 pre-trained model with 418M parameters \cite{fan2020beyond} to fine-tune into bi-directional Amharic-English NMT. For fine-tuning, we used max source \& target length of 128 per device train and validation, a batch size of 4, and 3 epochs. 

\subsection{Results and discussion}
In our experiments, we built Transformer models and fine-tuned M2M100\_418M, the multi-lingual pre-trained language model for bi-directional Amharic-English translation. All models were built on a regular (unnormalized) and normalized datasets to study the effect of Amharic homophone normalization in Amharic-English translation. Table \ref{tab:table3} shows the experimental results of Transformer and PLMs of M2M100 in both directions. 
\begin{table}[h!]
\centering
\caption{\label{tab:table3}
Experimental results}
\begin{tabular}{lcc}
\hline
\textbf{Models} & \textbf{Regular}& \textbf{Normalized}\\
\hline
Transformer (Amharic→English) & 14.78 & 16.26 \\
Transformer (English→Amharic) & 10.79 & 13.06 \\
M2M100 48M (Amharic→English) & 34.12 & \textbf{37.79} \\
M2M100 48M (English→Amharic) & 29.65 & \textbf{32.74} \\
\hline
\end{tabular}
\end{table}

As it can be observed from the results in Table \ref{tab:table3}, the multilingual translation model M2M-100 outperforms the Transformer-based models in both directions. When we compare our results with the attempts mentioned in Section \ref{related_work}, our research shows an improvement in parallel corpus size and BLEU scores using multilingual translation model. We can evaluate our models in the translation directions as follows.

\textbf{Amharic-English Translation:}
 For both Amharic-English and English-Amharic translations, the Transformer models did not work well, performed even less than some traditional methods. 
 The pre-trained transformer-based model performed better with 34.12 and 37.79 BLEU score for Amharic-English translation, on the regular and normalized data, respectively. This clearly shows that pre-trained based translation outperforms the baseline system.
 
\textbf{English-Amharic Translation:}
Among the works conducted for English-Amharic translation, the work by Teshome et al. \cite{teshome2015phoneme} reached a BLEU score of 37.53 using phoneme-based statistical MT, while our  pre-trained model showed a BLEU score of 29.65 and 32.74 on regular and normalized data, respectively. This phoneme-based work \cite{teshome2015phoneme} is built by converting the Amharic syllables to phoneme-based characters; however 1) the conversion is rule-based (unable to handle all exceptions), 2) Amharic homophones do not have constant Latin representations, and 3) we used new train and test sets. So, the results are not comparable. Another insight from this work is pre-trained model still needs improvement to translate from technologically favored languages to languages like Amharic which is morphologically rich.  
\begin{figure}[ht!]
    \centering
    \includegraphics[width=\linewidth]{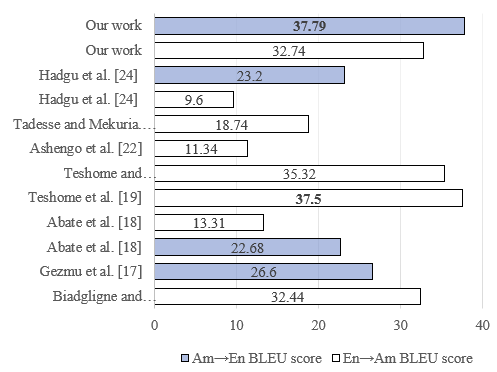}
   \caption{Reporting of our work result with other previous works}
    \label{fig:result}
\end{figure}

Our results also clearly show the effect of homophone character normalization in the performance of bi-directional Amharic-English translation. For both (Transformer and pre-trained) models, homophone character normalization increased the performance of NMT system. 
Accordingly, comparing our results with the previous attempts as shown in Figure \ref{fig:result}, we can confirm that our models demonstrated a 0.29\% BLEU score increase. This BLEU score improvement was made using our new training and testing sets. This is because, first, all the mentioned MT data is not available; second, the available data is not split into constant training and testing categories for future evaluation purposes. For the next time, our dataset will be available in train, test, and validation files. 

\section{Conclusion and future work}\label{conclusion}
In this paper, we presented our Amharic-English parallel corpus and bi-directional machine translation experiments. We gathered more than 888K parallel unique sentences and applied different pre-processing techniques. This is the first large-scale parallel data (more than 5 times bigger than the data in the previously conducted MT works) which can be used as a good benchmark for future machine translation research. The main gaps from the previous work were that they used small data and were not ready for future comparison (benchmark) by splitting the data into constant train, test, and validation sets. Our work will solve this issue and be used as a benchmark.
According to our result, we can conclude that the pre-trained models outperformed the baseline Transformer-based model. The morphological richness of the Amharic language and the size of the parallel dataset has a great impact on Amharic-English MT experiments. 
For the future, we will expand this work for more languages by including other Ethiopian low-resourced languages and also use data augmentation techniques. Additionally, we plan to explore the applicability of other pre-trained language models for Amharic-English translations. 
Our dataset of parallel Amharic-English sentence pairs, the models, and pre-processing scripts will be released in the GitHub repository\footnote{https://github.com/atnafuatx/EthioNMT-datasets}
\bibliographystyle{IEEEtran}
\bibliography{IEEEabrv,IEEEexample,conference_101719}
\end{document}